\documentclass[nohyperref]{article}

\usepackage{microtype}
\usepackage{graphicx}
\usepackage{subfigure}
\usepackage{booktabs} 

\usepackage{hyperref}

\usepackage[accepted]{icml2022}

\usepackage{amsmath}
\usepackage{amssymb}
\usepackage{mathtools}
\usepackage{amsthm}
\usepackage{hyperref}
\usepackage{url}
\usepackage[utf8]{inputenc} 
\usepackage[T1]{fontenc}    
\usepackage{booktabs}       
\usepackage{amsfonts}       
\usepackage{nicefrac}       
\usepackage{microtype}      
\usepackage{xcolor}         
\usepackage{amsmath}
\usepackage{wrapfig}
\usepackage{tikz}
\usepackage{pgfplots}
\usepackage{stmaryrd}
\usepackage{mathabx}
\newcommand\numberthis{\addtocounter{equation}{1}\tag{\theequation}}

\usepackage[capitalize,noabbrev]{cleveref}

\theoremstyle{plain}

\theoremstyle{definition}

\theoremstyle{remark}

\usepackage[textsize=tiny]{todonotes}

\icmltitlerunning{Learning to Infer the Structures of Network Games}

\begin{document}

\twocolumn[
\icmltitle{Learning to Infer Structures of Network Games}




\begin{icmlauthorlist}
\icmlauthor{Emanuele Rossi}{twitter,imperial}
\icmlauthor{Federico Monti}{twitter}
\icmlauthor{Yan Leng}{austin}
\icmlauthor{Michael M. Bronstein}{twitter,oxford}
\icmlauthor{Xiaowen Dong}{oxford}
\end{icmlauthorlist}

\icmlaffiliation{twitter}{Twitter, London, UK}
\icmlaffiliation{imperial}{Imperial College London, London, UK}
\icmlaffiliation{oxford}{University of Oxford, Oxford, UK}
\icmlaffiliation{austin}{The University of Texas at Austin, Austin, TX, USA}

\icmlcorrespondingauthor{Emanuele Rossi}{emanuele.rossi1909@gmail.com}

\icmlkeywords{Machine Learning, ICML}

\vskip 0.3in
]



\printAffiliationsAndNotice{}  

\begin{abstract}
Strategic interactions between a group of individuals or organisations can be modelled as games played on networks, where a player's payoff depends not only on their actions but also on those of their neighbours. 
Inferring the network structure from observed game outcomes (equilibrium actions) is an important problem with numerous potential applications in economics and social sciences. 
Existing methods mostly require the knowledge of the utility function associated with the game, which is often unrealistic to obtain in real-world scenarios.  
We adopt a transformer-like architecture which correctly accounts for the symmetries of the problem and learns a mapping from the equilibrium actions to the network structure of the game without explicit knowledge of the utility function. We test our method on three different types of network games using both synthetic and real-world data, and demonstrate its effectiveness in network structure inference and superior performance over existing methods. 
\end{abstract}

\section{Introduction}
Individuals or organisations cooperate with or compete against each other in a wide range of practical situations. Strategic interactions between individuals are often modeled as games played on networks \citep{jackson2014games}, where an individual's utility depends not only on their actions (and characteristics) but also on those of their neighbours. In a network game (and games in general), the utility function is of fundamental importance, as rational agents maximise their utilities when making real-life decisions. 
The current literature on network games has primarily focused on the scenarios where the utility function is predefined, and the structure of the network (represented by a graph) 
is known beforehand. 
However, in practical settings, while it is common to observe 
the actions of the players, 
the underlying interaction network often remains hidden or uncertain due to privacy reasons or the dynamic nature of interactions. Furthermore, the utility function is often unobservable. 
This makes it challenging to exploit network information and utility function for behavioural predictions and network-based interventions \citep{valente2012network}, e.g., marketing campaigns or information diffusion. 


In this paper, we focus on the problem of inferring the structure of the interaction network from observed equilibrium actions of a network game. A few recent studies have tackled similar problems \citep{irfan2011game,honorio2015learning,ghoshal2017a,ghoshal2017b,garg2016learning,garg2017local,barik2019provable,barik2020provable,leng2020learning, pmlr-v119-garg20b}; however, several major limitations remain. First, all these methods require the explicit knowledge of the utility function to infer the underlying network structure, which may be impractical to assume and may also change over time. The work of \citet{honorio2015learning} considers a more general hypothesis space of games for linear influence game, but they only focus on binary actions and linear payoffs. 
Second, the methodology in each of these studies has been designed for the specific game under consideration, thereby limiting its scope in handling a wide range of strategic interactions in real-world scenarios.

To address these limitations, we first summarise three common network games studied in the recent literature in a generic form, which is based on both individual and network factors that impact one's utility and the corresponding equilibrium actions. Despite the different nature of the games, it permits the relation between the equilibrium actions and network structure to be written in a unified manner. This motivates us to propose a data-driven model where we learn the functional mapping between the equilibrium actions and network structure {\em without} explicit knowledge about the utility function of the game. Our model is based on an encoder-decoder approach where the encoder is a transformer architecture and the decoder can be chosen flexibly, and is trained using pairs of actions and network structures. Once trained, the model can be deployed to infer the network structure from only observed actions. Synthetic and real-world experiments demonstrate the superiority of our method against several state-of-the-art approaches.


\paragraph{Main contributions.}
First, we compare three network games using a unified parameterisation, which helps reveal the different nature of these games and interpret the strategic interactions they represent. Second, to our knowledge, our framework is one of the first that is able to infer the network structure behind the games without explicit knowledge about the utility function. This capability is important in real-world scenarios where the nature of the interactions remains hidden or may even evolve over time. Finally, our work contributes to the emerging field of data-driven structural inference by proposing a model based on a novel architecture which adopts a permutation-invariant transformer. Overall, our work contributes to the inference of strategic decision making in various network settings.


\section{Related Work}

{\em Network games}, a class of problems in game theory, have been studied extensively in computer science and economics. The majority of works in the network game literature study the characteristics of  games on a known and static graph \citep{ballester2006s, bramoulle2014strategic}. 
While these studies are useful in understanding collective actions and designing interventions \citep{galeotti2017targeting}, it is increasingly acknowledged that networks are difficult to obtain in practice. 
Furthermore, the utility function associated with the game is usually unknown as well. 
%
We are interested in the {\em inverse problem} of inferring the network structure based on observed actions. 
This inverse setting is related to {\em graph} or {\em network inference}, a problem that has attracted interest in statistics \citep{Koller09, Friedman08}, physics \citep{GomezRodriguez_2010, Gomez-Rodriguez20011}, signal processing \citep{Mateos19,Dong19}. Our study differs from these works in accounting for the strategic interactions and the game theoretical framework underlying the observed data. 

Deep learning models have been recently proposed for latent graph inference in a number of settings.
\citet{kipf2018neural} proposes a graph neural network (GNN) model to infer the interactions of an underlying dynamical system from observational data. Differently from this work, their model is trained on predictions of the future state of the system, where there is lack of validation for the learned network interactions. Similarly, in latent graph learning \citep{DBLP:conf/miccai/CosmoKANB20, DBLP:journals/corr/abs-2003-13620,10.1145/3326362}, the graph is learned jointly with a downstream task, conversely to our scenario where the structure itself is the learning objective.
Methods for link prediction \citep{zhang2018link} predict edges in a graph, but they typically require part of the true links to be provided as input (and only predict the missing ones), whereas in our scenario we are interested in inferring networks without observing any link in the test data.
Among the few works where the network structure is the learning objective, \citet{yu2019dag} and \citet{zheng2020learning} propose to infers a DAG from the observed actions. However, these approaches are limited to predicting acyclic graphs, whereas the graphs we are interested in are often cyclic. 
The method of \citet{belilovsky2017learning} is the mostly related to ours, since they propose a supervised model to infer an undirected graphical model from observed covariates using a series of dilated convolutions. However, their model is not permutation-equivariant w.r.t. the order of nodes and the number of layers depends on the number of nodes. Both issues cause the statistical efficiency of the model to scale poorly with the size of the graph. Different from the above studies, our framework aims to learn game-theoretical relationships in a supervised manner, while leveraging the structural symmetries of this 
problem. 

Finally, there has been a recent stream of literature in learning network games from actions of players \citep{irfan2011game,honorio2015learning,ghoshal2017a,ghoshal2017b,garg2016learning,garg2017local,barik2019provable,barik2020provable,leng2020learning,garg2020predicting}. Most of these methods focus on either a binary or a finite discrete action space. 
For continuous actions, \citet{leng2020learning} formulate an optimisation problem to learn the structure and marginal benefits of linear quadratic games, 
while \citet{barik2019provable} aim at inferring the network structure from an action-conforming graphical game. 
\citet{garg2020predicting} learn a mapping from observed actions to interaction structures where individuals repeatedly update their strategies based on a weighted aggregate of other players' choices. 
Our work differs from existing methods in the literature in that it does not assume a specific game-theoretic structure (e.g., utility functions). Instead, we build a transformer-like model that learns a mapping from the equilibrium actions to the network structure of the games without explicit knowledge of the utility functions.

\section{Setting}

In this section, we start by analysing three commonly studied network games (Sections \ref{sec:game} and \ref{sec:idiosyncratic}). Based on the specific utility function of these games, we establish a generic relationship between the equilibrium actions and network structure (Section \ref{sec:action}). This eventually motivates the proposed framework that learns network structure without any knowledge of the utility function\footnote{The utility function is used only in analysing three classical games in a universal framework as well as generating data in synthetic experiments. It is not used by the proposed method itself.} (Section \ref{sec:approach}).

\subsection{Continuous-action network games}
\label{sec:game}

\begin{figure*}[t]
    \centering
    \includegraphics[width=\linewidth]{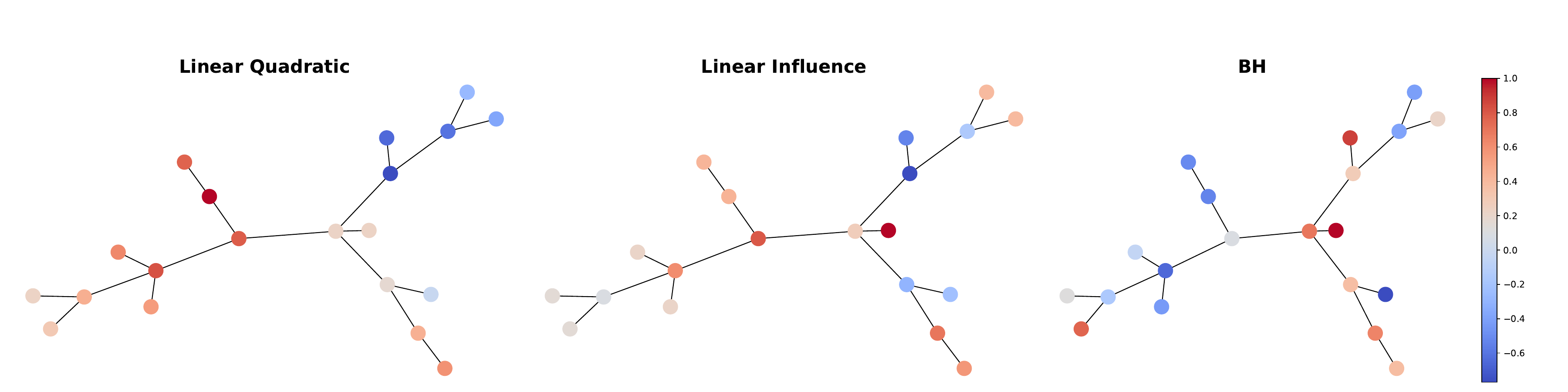}\vspace{-0.4cm}
    \caption{Example actions for different types of games on a Barabási-Albert graph. Actions are normalised to be in $[-1, 1]$ and are displayed as colors on nodes.}
    \vspace{-0.4cm}
    \label{fig:game_examples}
\end{figure*}

Consider a network of $N$ individuals represented by a weighted and undirected graph $\mathcal{G}(\mathcal{V},\mathcal{E})$, where $\mathcal{V}$ and $\mathcal{E}$ denote the node and edge sets, respectively. For any pair of individuals $i$ and $j$, $w_{ij}$ = $w_{ji} > 0$ if $(i,j) \in \mathcal{E}$ and $w_{ij} = w_{ji} = 0$ otherwise, where $w_{ij}$ is the $ij$-th entry of the adjacency matrix $\mathbf{W}$. In this work, we assume the graph $\mathcal{G}$ is connected, and consider the normalised adjacency matrix $\mathbf{A} = \mathbf{D}^{-\frac{1}{2}} \mathbf{W} \mathbf{D}^{-\frac{1}{2}}$ where $\mathbf{D} = \text{diag}(\mathbf{W}\mathbf{1})$, where $\mathbf{1}$ is an $N$-dimensional vector of ones.
In a network game, the payoff $u_i$ of an individual $i$ depends on their action $x_i$ as well as the actions of neighbours $j \in \mathcal{N}_i$. 
We consider three commonly studied network games: linear quadratic games \citep{ballester2006s}, a variation of the linear influence games developed in \citet{irfan2011game}, and the graphical game studied in \citet{barik2019provable} (which we will refer to as Barik-Honorio or BH graphical game).




\noindent \textbf{Linear quadratic games.}
   {Linear quadratic games are widely studied in the economics literature \citep{jackson2014games, ballester2006s}. 
   In this game, a player $i$ chooses their action by maximising the following utility function:} 
   \begin{equation}
      \max_{\{x_i\}} u_i = b_i x_i -\frac{1}{2} x_i^2 + \beta \sum_{j \in \mathcal{N}_i} a_{ij} x_i x_j   , 
       \label{eq:util_fixed_cost}
   \end{equation}
   where $b_i$ represents the marginal benefit of $i$ by taking action $x_i$ and $\beta$ is the strength of dependencies between actions of neighbours in the network, respectively. 
   Note that this utility function can also be thought of as a second-order approximation to non-linear utility functions of more complex games.
   The pure-strategy Nash equilibrium (PSNE) of this game is 
   \begin{equation}
   \mathbf{x}^* = \big( \mathbf{I} - \beta \mathbf{A} \big)^{-1} \mathbf{b},
   \label{eq:equi_fixed_cost}
   \end{equation}
   where $\mathbf{x}^*$ and $\mathbf{b}$ are $N$-dimensional vectors  collecting actions and marginal benefits for all individuals, and $\mathbf{I}$ is the $N \times N$ identity matrix. Under the assumption that $|\beta|<1$, the matrix inverse is guaranteed to exist as the spectral radius of $\mathbf{A}$ is 1. Furthermore, when $\beta>0$, the game corresponds to a strategic complement relationship (i.e., intuitively, the incentive of a player to take a higher action is increasing in the number of their neighbours also taking a higher action); when $\beta<0$, it corresponds to strategic substitute (i.e., intuitively, the incentive of a player to take a higher action is decreasing in the number of their neighbours also taking a higher action).

\noindent \textbf{Linear influence games.} Inspired by the threshold model \citep{granovetter1978threshold}, \citet{irfan2011game} proposed the linear influence games, where an individual chooses the action that maximises the following utility function\footnote{
The actions are discrete in the originally proposed game. We adapt the game to a continuous setting.}:
    \begin{equation}
        \max_{\{x_i\}} u_i =  \sum_{j \in \mathcal{N}_i} a_{ij} x_i x_j - b_i x_i, 
    \label{eq:util_lig}
    \end{equation}
    where $b_i$ can be understood as a threshold parameter for $i$'s level of tolerance for negative effects. 
   Under the assumption that $\mathbf{A}$ is invertible, the PSNE satisfies the following condition: 
    \begin{equation}
        \mathbf{x}^* = \mathbf{A}^{-1} \mathbf{b}. 
    \label{eq:equi_lig}
    \end{equation}
    
\noindent \textbf{BH graphical game.} 
Finally, in the specific graphical game introduced in  \citet{barik2019provable}, an individual maximises the following utility function: 
\begin{equation}
    \max_{\{x_i\}} u_i = - \Big\| x_i - \sum_{j \in \mathcal{V}}a_{ij}x_j\Big\|_2. 
\label{eq:util_graphical_game}
\end{equation}
This utility can be used to model games where an individual prefers to conform to the social norm (actions of their neighbours), since their own utility decreases as their action deviates from those of their neighbours. 
The PSNE for this game satisfies the following condition:
\begin{equation}
    \mathbf{x}^{*} = \mathbf{A} \mathbf{x}^{*}. 
\label{eq:equi_graphical_game}
\end{equation}
This suggests that $\mathbf{x}^{*}$ is the eigenvector $\mathbf{u}_1$ of $\mathbf{A}$ which is associated with the largest eigenvalue (which is 1). 
We consider actions from the set of $\epsilon$-PSNE \citep{barik2019provable}, which are obtained by adding noise independently per player. The observed actions $\mathbf{x}$ are: $\mathbf{x}= \mathbf{x}^* + \mathbf{e}$,
where $\mathbf{e}$ is Gaussian noise.

%
\begin{table}
\caption{Parameterisation of three network games.}
\vspace{0.2cm}
\label{tab:summary_game}
\centering
\begin{tabular}{|lll|}
\hline
                                        & $\mathcal{F} (\mathbf{A})$                           & $\mathcal{H} (\mathbf{b})$    \\ \hline
Linear quadratic   & $(\mathbf{I} - \beta \mathbf{A})^{-1}$    & $\mathbf{b}$       \\ \hline
Linear influence                    &  $\mathbf{A}^{-1}$                        & $\mathbf{b}$       \\ \hline
Barik-Honorio                          & $\mathbf{u}_1$                    & 1           \\ \hline
\end{tabular}
\end{table}

Example actions for the three types of games on a Barabási-Albert graph have been illustrated in Fig.~\ref{fig:game_examples}.
By investigating
Eq.~\eqref{eq:equi_fixed_cost}, \eqref{eq:equi_lig},
and \eqref{eq:equi_graphical_game}, we can write the condition for the equilibrium actions $\mathbf{x}^*$ in a generic form\footnote{We tacitly assume that $|\beta|<1$ and $\mathbf{A}$ is invertible.} (see Table~\ref{tab:summary_game}): 
\begin{equation}
    \mathbf{x}^* = \mathcal{F} (\mathbf{A}) \mathcal{H} (\mathbf{b}), 
\label{eq:generalization}
\end{equation}
where $\mathcal{F}(\mathbf{A})$ is a function of the network structure and $\mathcal{H}(\mathbf{b})$ is a function of additional parameters (if any) associated with the game. 
That is, $\mathcal{F}(\mathbf{A})$ accounts for the influence from the actions of one's neighbours in the network. 
Conversely, $\mathcal{H} (\mathbf{b})$ is only affected by one's idiosyncratic (individual) characteristics. 
\subsection{Modeling of Individual Idiosyncratic Characteristics}
\label{sec:idiosyncratic}

Under the linear quadratic or linear influence games, the parameter $\mathbf{b}$ captures the marginal benefits or tolerance levels as idiosyncratic characteristics of players in the corresponding games. In the presence of the homophily effect \citep{mcpherson2001birds}, we may assume that this parameter is associated with the network structure. To this end, we propose to model $\mathbf{b}$ as follows:
\begin{equation}
   \mathbf{b} \sim \mathcal{N}(\mathbf{0}, \mathbf{L}^\dagger_\alpha),\quad\quad \mathbf{L_\alpha} = (1-\alpha) \mathbf{I} + \alpha \mathbf{L}
   \label{eq:b}
\end{equation}
where $\mathbf{L} = \mathbf{I} - \mathbf{A}$ is the normalised graph Laplacian matrix, and $^\dagger$ represents pseudoinverse in case the matrix is not invertible (this happens when $\alpha=1$ as $\mathbf{L}$ has a smallest eigenvalue of 0). The parameter $\alpha \in [0,1]$ controls the relation of the individual idiosyncratic characteristics to the network structure.
The two corner cases $\alpha=0$ (for which we have $\mathbf{L}_0 = \mathbf{I}$) and $\alpha=1$ (when $\mathbf{L}_1=\mathbf{L}$) correspond to independent idiosyncratic characteristics and homophilous idiosyncratic characteristics (individuals with similar characteristics tend to be connected), respectively. By varying $\alpha$ from 0 to 1 we can achieve increasing levels of homophily or smoothness (see Section~\ref{sec:action}) of $\mathbf{b}$ on the graph.

\subsection{Analysis of Equilibrium Actions}
\label{sec:action}
With the conditions for equilibrium actions in Section~\ref{sec:game} and the modeling of individual idiosyncratic characteristics in Section~\ref{sec:idiosyncratic}, we can analyse explicitly the characteristics of these actions.

\begin{figure}
\includegraphics[width=1\linewidth, height = 4.5cm]{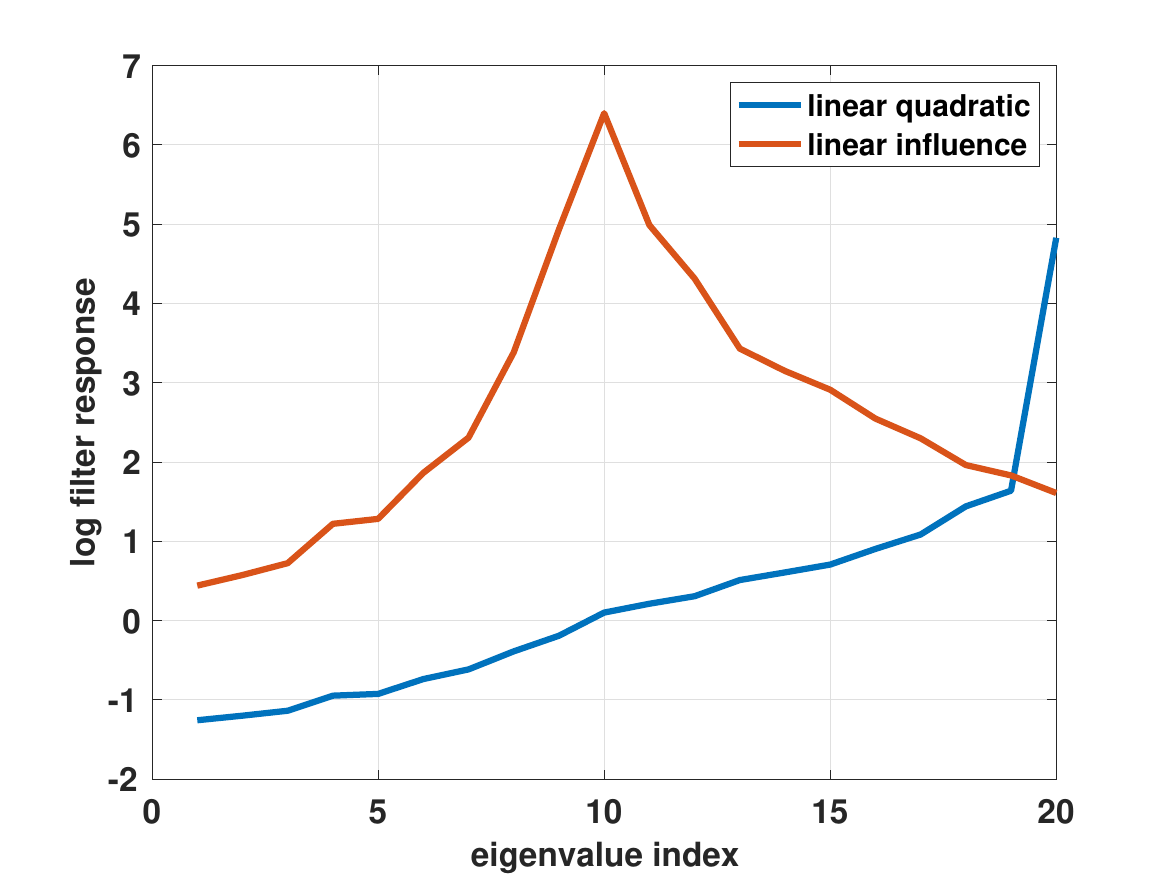}
\caption{Interpreting actions of linear quadratic/influence games: the equilibrium actions for the two games are dominated by different sets of eigenvectors.}
\vspace{-0.5cm}
\label{fig:cov_filter}
\end{figure}

\noindent \textbf{Linear quadratic games.}
Assuming $\mathbf{b}$ of the form~\eqref{eq:b} and using 
Eq.~\eqref{eq:equi_fixed_cost} and 
$\mathbf{L} = \mathbf{I} - \mathbf{A}$, we have that the equilibrium actions $\mathbf{x}^*$ follow a multivariate Gaussian distribution:

\vspace{-0.5cm}\begin{align}
\begin{split}
   \mathbf{x}^* &\sim \mathcal{N}\Big(\mathbf{0}, \big( \mathbf{I} - \beta \mathbf{A} \big)^{-1} \big( \mathbf{I} - \alpha \mathbf{A} \big)^\dagger \big( \mathbf{I} - \beta \mathbf{A} \big)^{-1}\Big) \\ &= \mathcal{N}\Big(\mathbf{0}, \mathbf{U} [(\mathbf{I} - \beta \mathbf{\Lambda})^2 (\mathbf{I} - \alpha \mathbf{\Lambda})]^\dagger \mathbf{U}^\top\Big),
   \label{eq:act_fixed_cost}
\end{split}
\end{align}

with the eigendecomposition $\mathbf{A} = \mathbf{U} \mathbf{\Lambda} \mathbf{U}^\top$. 
Eq.~\eqref{eq:act_fixed_cost} illustrates the relationship between the actions $\mathbf{x}$ and the network structure $\mathbf{A}$ and motivates the learning framework proposed in the next section. Furthermore, the covariance in Eq.~\eqref{eq:act_fixed_cost} may be interpreted as a graph filter whose frequency response 
$\tfrac{1}{(1-\beta \lambda)^2(1 - \alpha \lambda)}$ 
may shed light on the behaviour of the actions. For $\beta \rightarrow 1$  and $\alpha \rightarrow 1$, the action vector $\mathbf{x}$ tends to behave like the leading eigenvectors of $\mathbf{A}$ (`low frequency'), which are smooth on the graph\footnote{By `smoothness' here we mean the Dirichlet energy of the features, i.e., $\mathrm{trace}(\mathbf{X}^\top \mathbf{L} \mathbf{X})$.}. An example of the filter response (with $\beta=0.8$ and $\alpha=0.8$) applied on the eigenvalues of an instance of a 20-node Erdős-Rényi graph is shown in Fig.~\ref{fig:cov_filter} (blue).
This shows that in this case the actions of linear-quadratic games are dominated by the leading eigenvector of A which is smooth.

\noindent \textbf{Linear influence games.}
Similarly, from Eq.~\eqref{eq:equi_lig} and~\eqref{eq:b}, we have that the equilibrium actions $\mathbf{x}$ follow a multivariate Gaussian distribution:
\begin{align}
\begin{split}
   \mathbf{x}^* &\sim \mathcal{N}\Big(\mathbf{0}, \mathbf{A}^{-1} \big( \mathbf{I} - \alpha \mathbf{A} \big)^\dagger \mathbf{A}^{-1}\Big) \\&= \mathcal{N}\Big(\mathbf{0}, \mathbf{U} [\mathbf{\Lambda}^2 (\mathbf{I} - \alpha \mathbf{\Lambda})]^\dagger \mathbf{U}^\top\Big).
   \label{eq:act_lig}
\end{split}
\end{align}
Interpreting Eq.~\eqref{eq:act_lig} as a spectral filter of the form $\tfrac{1}{\lambda^2 (1-\alpha \lambda)}$ we can also conclude that an $\alpha \rightarrow 1$ tends to lead to smoother actions on the graph. However, the exact behaviour of actions in this case depends on the magnitude of the eigenvalue of $\mathbf{A}$ closest to 0. Given that the spectrum of $\mathbf{A}$ lies in the range of $[-1,1]$, the actions are likely to behave like mid-spectrum eigenvectors, which are not necessarily smooth signals on the graph. Similarly, this can be seen from the filter response, under $\alpha=0.8$, shown in Fig.~\ref{fig:cov_filter} (red). 
This shows that in linear influence games the actions are dominated by mid-spectrum eigenvectors which are not necessarily smooth.

\noindent \textbf{BH graphical game.}
We see from Eq.~\eqref{eq:equi_graphical_game} that the equilibrium actions correspond to the largest eigenvector $\mathbf{u}_1$ of $\mathbf{A}$. Although in this setting the observed actions are $\epsilon$-PSNE, they would still tend to be smooth on the graph.

In summary, our analysis in this section shows that the smoothness of the equilibrium actions on the graph depends on $\beta$ and $\alpha$ in the linear quadratic game. The actions in linear influence game are likely to be nonsmooth, whereas those in the BH graphical game  are likely to be  smooth. We empirically validate this analysis in Section~\ref{sec:synthetic_data}. Regardless of the smoothness, the relationship between the equilibrium actions and the network structure demonstrated in this section motivates us to propose a learning framework in the next section to infer the network structure from the observed actions.

\section{Proposed Approach} \label{sec:approach}
Motivated by the analysis in Section \ref{sec:action}, we propose a model that learns a direct mapping from the observed actions to the network structure. Such a model is agnostic to the utility function of the game and avoids strong assumptions on it (as long as it conforms to the broad class of games whose equilibrium actions can be parameterised by Eq.~\eqref{eq:generalization}). 
Specifically, we consider the scenario where social network and decision data exist for a small population. 
The objective is to learn the mapping from decisions to the network structure on this small population such that it can be used to infer a large-scale unobserved network.
For example, for cost-effective data collection, government, public agencies, and researchers can collect social network data on a small population (by asking individuals to nominate their friends) and then use the proposed method to learn the network interactions for a larger population. 

In our setting, we assume to have a training set $\mathcal{D}$ of action-graph pairs $(\mathbf{X}^{(\ell)}, \mathbf{A}^{(\ell)})$ coming from games with the same (but unknown) utility function. 
For each $\ell$, the model takes as input an $N \times K$ matrix $\mathbf{X}^{(\ell)}$ containing the actions of $N$ players over $K$ independent games as columns, and outputs a predicted $N \times N$ adjacency matrix $\hat{\mathbf{A}}^{(\ell)} = g_\Theta(\mathbf{X}^{(\ell)})$ for the network game (we drop $\ell$ from now on for simplicity).
Once trained, our model can then infer the network structure corresponding to previously unseen actions, as long as they are generated following a similar utility function. Moreover, this framework makes our model generalisable to learning networks with the number of nodes different from what was observed during training.

The model parameters are learned such that they minimise the
  cross-entropy between the binary ground truth adjacency matrix $\mathbf{A}$ and the predicted continuous one $\hat{\mathbf{A}}$. 
%
Binary cross-entropy is a standard loss function for link prediction tasks using graph neural networks \citep{NIPS2017_5dd9db5e}, of which our problem is an instance. We also experimented with weighting the loss based on the proportion of edges in the graph, but it did not change the performance of the model.


Our model follows an encoder-decoder architecture $g_\Theta(\mathbf{X}) = \mathrm{dec}(\mathrm{enc}(\mathbf{X}) )$, which is a standard solution for link-prediction problems in graph neural networks \citep{kipf2016variational} since the resulting number of parameters of the model is independent on the size of the graph, allowing for more statistical efficiency, and for the same model to work on graphs of different sizes.
The {\em encoder} outputs an $N\times K \times F$ tensor $\mathbf{Z} = \mathrm{enc}(\mathbf{X})$, mapping each node $i$ in game $k$ to an $F$-dimensional latent embedding $\mathbf{z}_{ik}$. The {\em decoder} outputs the predicted $N\times N$ adjacency matrix $\hat{\mathbf{A}} = \mathrm{dec}(\mathbf{Z})$, where the score $\hat{a}_{ij}$ for each edge $(i,j)$ is computed using the $K\times F$-dimensional embeddings $\mathbf{Z}_{i},\mathbf{Z}_{j}$ of the respective nodes. 

Given that the ordering of nodes in the graph is arbitrary, the model has to be a {\em permutation-equivariant} \citep{gdlbook} function over the set of nodes: $g_\Theta(\mathbf{P}_1 \mathbf{X})=\mathbf{P}_1 g_\Theta(\mathbf{X})\mathbf{P}_1^\top$, where $\mathbf{P}_1$  is an $N\times N$ permutation matrix interpreted as reordering of the nodes of the graph. Additionally, in the case where there is no correspondence between game $k$ in one graph and game $k$ in another one (e.g., actions corresponding to user rating where however users in different graphs have rated different items), the model should also be {\em permutation-invariant} over the set of games, i.e., $g_\Theta(\mathbf{X} \mathbf{P}_2)=g_\Theta(\mathbf{X})$, where $\mathbf{P}_2 $ is another $K\times K$ permutation matrix on the games. Overall, this combined symmetry condition can be written as $g_\Theta(\mathbf{P}_1 \mathbf{X} \mathbf{P}_2)= \mathbf{P}_1 g_\Theta(\mathbf{X})\mathbf{P}_1^\top$.

\vspace{-0.2cm} 

\paragraph{Encoder} 
The input to our model is a variable-length set (of actions), which excludes using a multi-layer perceptron (since it would not be able to handle the variable length) and sequence models such as LSTMs \citep{HochSchm97} or GRUs \citep{cho-etal-2014-learning} (since they treat the input as an ordered sequence, while it is an un-ordered set).
Moreover, since the ground truth graph between the players is not known a priori, but it is however important for the model to exchange information between players (as the values of agents' actions are meaningless if not compared to the ones of the others), the encoder needs to perform message passing on the fully connected graph of players.  Therefore, we propose a Transformer-like \citep{vaswani2017attention} encoder (which we refer to as  Network Game Transformer or \textit{NuGgeT}). \textit{NuGgeT} processes a \textit{set} of games happening on a given network and outputs for each node a \textit{set} of 
$K$ different game-specific embeddings (one for each game that happens on the same network). 
In this architecture, the value of action $x_{ik}$ of player $i$ in game $k$ is never mixed with the actions of other players in a different game $k'$, if not for the computation of an aggregated attention score $\alpha_{ij}$ that captures the overall similarity between players $i$ and $j$. 
Intuitively, in settings where we are given the outcomes of multiple games played on multiple graphs,  there is no correspondence between such games. Thus, the value of an action does not bring any information about the role of a node in a graph and the only useful information that needs to be exchanged across games is the similarity of the nodes. 

Specifically, for each node $i=1,\hdots, N$ and game $k=1,\hdots, K$, the embedding $\mathbf{z}_{ik}$ is computed as follows:
\begin{align*}
    \mathbf{y}_{ik}      &= \text{ReLU}(x_{ik} \boldsymbol{w} + \mathbf{b})\\
    \alpha_{ij}^{(h)} &= \text{softmax}_{j}\Big( \mathop{\bigboxvoid}_{k=1}^K  {\mathbf{y}}_{ik}^\top \mathbf{W}_Q^{(h)}\mathbf{W}_K^{(h)}{\mathbf{y}}_{jk} \Big)\\
    \mathbf{z}_{ik} &= \phi\Big({\mathbf{y}}_{ik},  \sum_{j=1}^N \alpha_{ij}^{(1)} {\mathbf{y}}_{jk},  \ldots ,  \sum_{j=1}^N \alpha_{ij}^{(H)} {\mathbf{y}}_{jk} \Big) \numberthis \label{eq:encoder} 
\end{align*}
where $\mathop{\bigboxvoid}$ denotes a general permutation-invariant aggregation operator (e.g $\max$, mean $\frac{1}{K}\sum_{k=1}^K$ or sum $\sum_{k=1}^K$), $\psi$ is some learnable function,  $\boldsymbol{w} \in \mathbb{R}^F$, $\mathbf{b} \in \mathbb{R}^F$, 
$\mathbf{W}^{(h)}_K \in \mathbb{R}^{F\times F'}$, and $\mathbf{W}^{(h)}_Q \in \mathbb{R}^{F'\times F}$ are learnable parameters, and 
$h=1,\hdots, H$ denotes the attention heads. 
For a generic node $i$ in game $k$, NuGgeT first expands $i$'s action $x_{ik}$ into a vector $\mathbf{y}_{ik}$ of $F$ features, 
then computes $H$ attention scores (one for each head) for each pair of nodes $(i, j)$ via multiplicative attention on the expanded actions $\mathbf{y}_{.k}$\footnote{For a given player $i$, using the expanded actions $\mathbf{y}_{.k}$ rather than input scalars $x_{.k}$ allows the attention mechanism to produce attention scores which are not necessarily linearly dependent on the value of neighbours' action $x_{jk}$, thus producing richer attention scores (see Fig.~\ref{fig:attention_scores} in Appendix).} (aggregating the unnormalised scores across games), and finally refines vector $\mathbf{y}_{ik}$ via a function $\psi$ that processes the $H$ aggregated representation of the neighbours obtained from the attention mechanism.


\paragraph{Decoder} In the case of directed graphs, there is no specific requirement for the decoder. If the graph is undirected, and therefore the adjacency matrix is symmetric, the decoder should be symmetric w.r.t. 
pairs of nodes (i.e., $(i,j)$ and $(j,i)$ treated the same way). 
Additionally, we impose  invariance w.r.t. ordering of the games.  
\textit{NuGgeT}'s decoder computes the predicted adjacency $\hat{a}_{ij}$
by aggregating the game-specific embeddings computed in~(\ref{eq:encoder}) for a pair of nodes $(i, j)$ using a general permutation-invariant aggregation operator $\mathop{\bigboxdot}$, which is then passed through a learnable function $\psi$:
\begin{align}
    \hat{a}_{ij} \psi\Big(\mathop{\bigboxdot}_{k=1}^{K} \mathbf{z}_{ik} \odot \mathbf{z}_{jk}\Big).
\end{align}

 Here $\odot$ denotes element-wise product, 
 whose use ensures symmetry w.r.t. node pairs (since $\mathbf{z}_{ik} \odot \mathbf{z}_{jk} = \mathbf{z}_{jk} \odot \mathbf{z}_{ik}$), while
 the permutation invariant operator over $k$ ensures invariance to the ordering of games.  
 %
 We empirically observe this approach to work better compared to simpler permutation-invariant functions such as the dot product of the concatenation of the embeddings from multiple games. 
 In the SM we prove that \textit{NuGgeT} satisfies the symmetry conditions outlined above.



\section{Experiments} \label{sec:experiments}
Our implementation of \textit{NugGeT} uses the sum $\sum_{k=1}^K$ as the permutation-invariant functions $\mathop{\bigboxvoid}$ and $\mathop{\bigboxdot}$, and two different 2-layer MLPs for $\phi$ and $\psi$. We use the Adam optimiser \citep{DBLP:journals/corr/KingmaB14} with a learning rate of $0.001$, a batch size of $100$ and a patience of $50$ epochs. We did not perform any particular hyperparameter tuning for our method, since we found it to be quite robust to the choice of hyperparameters and perform well with standard choices (see Appendix).  In all our experiments, we report the mean and the standard error of the mean over the test graphs. Note that we ignore diagonal elements of the adjacency matrix both for training and evaluation. Since they are always zero, the model could easily memorise them, influencing the metrics. We use an AWS p3.16xlarge machine with 8 GPUs. While the training of our model takes between 5 and 10 minutes on a single GPU, the whole set of experiments conducted in the paper necessitate roughly 4 days of GPU time.

\subsection{Synthetic Data} \label{sec:synthetic_data}

\begin{figure*}[h!btp]
    \centering\vspace{-12mm}
    \includegraphics[width=1\linewidth]{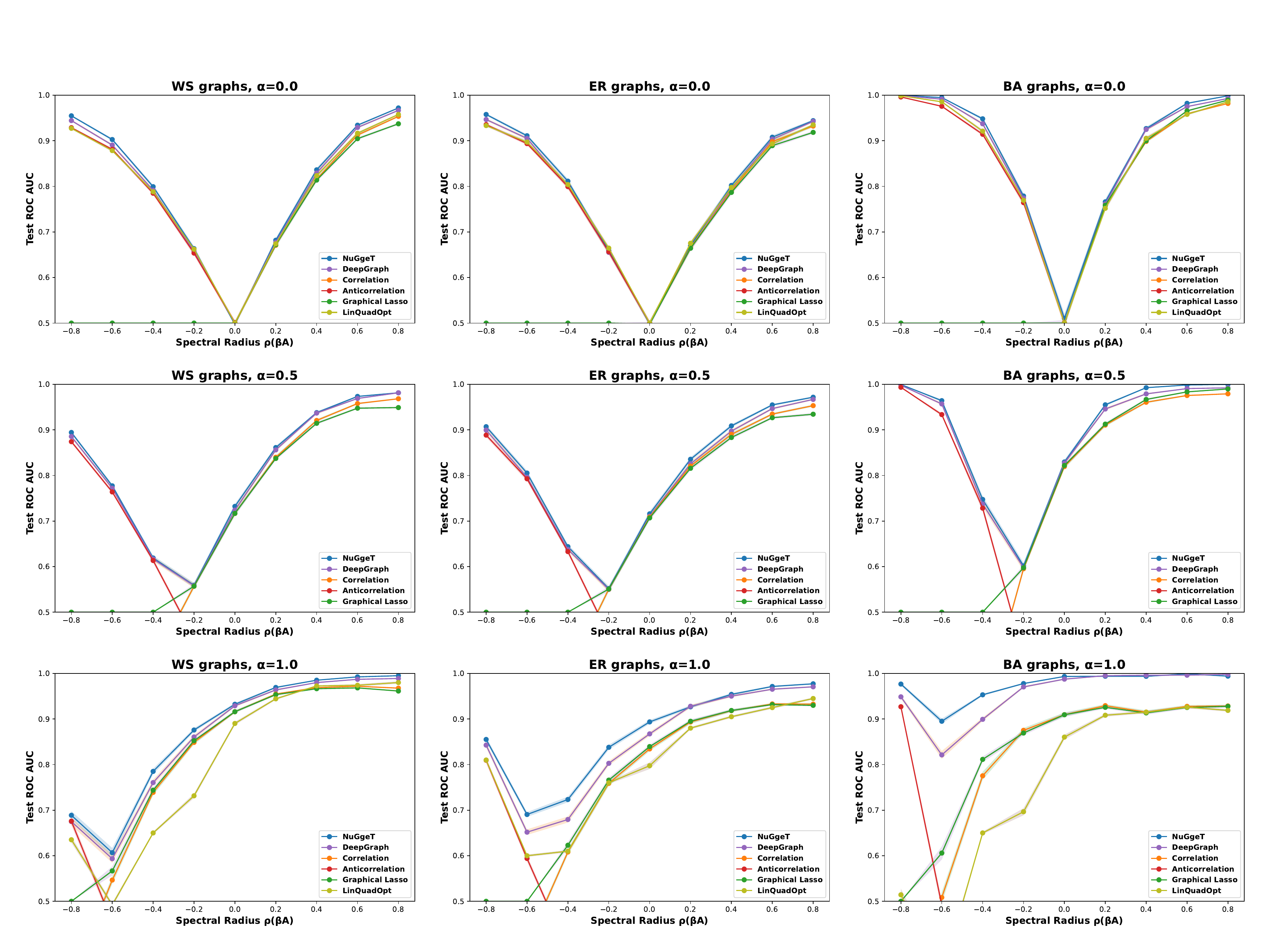}\vspace{-10mm}
    \caption{Results for linear quadratic games with varying $\alpha$ and spectral radius $\rho(\beta A)$.}
    \label{fig:linear_quadratic}
    \vspace{-0.5cm}
\end{figure*}

\paragraph{Data Generation}
We follow the setup in \citet{leng2020learning} for generating the synthetic graphs using three different random graph models: Erdős-Rényi (ER), Watts-Strogatz (WS), and Barabási-Albert (BA). More details and exact parameters for the synthetic data are provided in the Appendix. All the graphs have $N = 20$ vertices in our experiments. For each type of graphs above, we simulate equilibrium actions for linear quadratic, linear influence, and BH graphical games using their respective utility function. For linear quadratic games, once the graphs are constructed, we compute $\beta > 0$ such that the spectral radius $\rho(\beta A)$ varies between $0$ and $1$. For linear influence games, when the adjacency matrix is not invertible we take its pseudoinverse. For BH games, we set $\text{std}(\mathbf{e})=1$ and ensure the resulting actions are $\epsilon$-PSNE with $\epsilon=0.2$.
The generated actions together with the ground truth network structure are used to train the model.
We use 850 graphs in the training set, 50 graphs for validation and 100 graphs for testing and verify that there is no overlap between them.
It is important to notice that while $\alpha$, $\beta$ and $\epsilon$ are used to generate the synthetic data, they are not used or known by the model.

\paragraph{Baselines} We compare with the following general baselines: \textit{Correlation}, \textit{Anticorrelation}, \textit{Graphical Lasso} \citep{friedman_sparse_2008} and \textit{DeepGraph} \citep{belilovsky2017learning}. We also compare to game-specific baselines: \textit{LinQuadOpt (independent)} \citep{leng2020learning} and \textit{LinQuadOpt (homophilous)} \citep{leng2020learning} for linear quadratic games, and \textit{BlockRegression} \citep{barik2019provable} for BH Graphical Games. A more detailed description of the baselines and how they are tuned is provided in the Appendix.

\paragraph{Results} Results for Linear Quadratic Games are reported in Fig.~\ref{fig:linear_quadratic}. Columns are different types of graphs (ER, WS, BA), rows are different values of $\alpha$ that controls smoothness of the marginal benefits ($\alpha=0$, $\alpha=0.5$ and $\alpha=1$), while the x-axis represents the spectral radius $\rho(\beta A)$. \textit{NuGgeT} is on par or superior to other methods in all scenarios, with \textit{DeepGraph} being the runner-up competitor. 
We observe larger performance gap with $\alpha=1$, i.e., when the distribution of marginal benefits largely depends on graph structure. Interestingly, it can be observed that in this case \textit{NuGgeT} and \textit{DeepGraph} perform well in both cases of strategic complements ($\rho(\beta A) > 0$, neighbours take similar actions) and strategic substitutes ($\rho(\beta A) < 0$, neighbours take opposite actions), whereas other baselines only perform well in one of the two cases. This is due to the former two methods learning directly a mapping
between actions and graph structure that may correspond to different characteristics of the equilibrium actions.

\begin{figure*}[h!btp]
    \vspace{-0.6cm}
    \centering
    \includegraphics[width=1\linewidth]{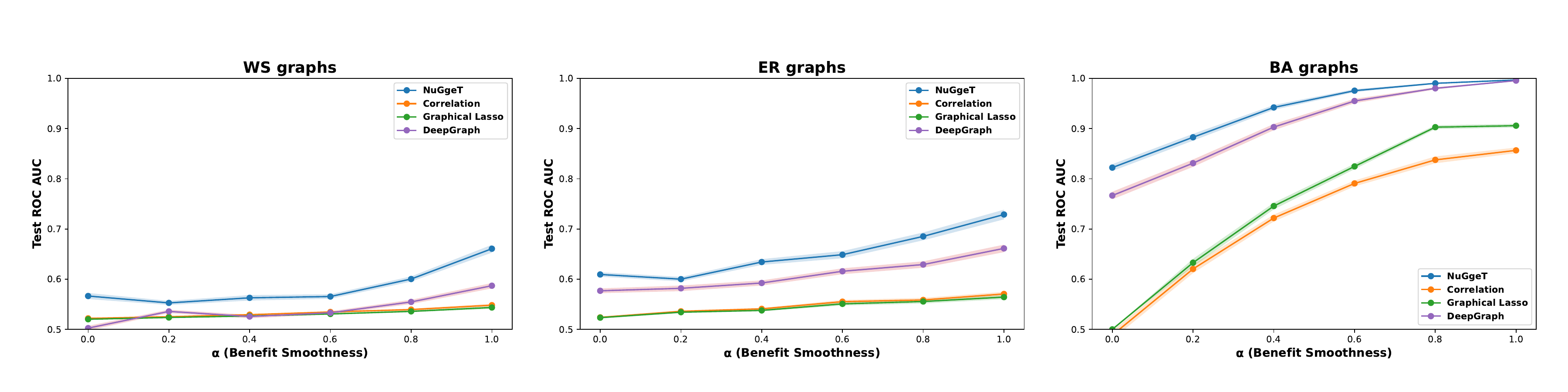}\vspace{-3mm}
    \vspace{-0.4cm}
    \caption{Results on linear influence games when varying the smoothness $\alpha$ of the marginal benefits.}
    \label{fig:linear_influence}
    \vspace{-1mm}
\end{figure*}

For Linear Influence Games, we report the results in Fig.~\ref{fig:linear_influence}. Each plot corresponds to a different type of graph, and the x-axis represents the benefit smoothness $\alpha$. As expected, all methods improve their performance as $\alpha$ grows, since the actions generally become smoother over the graph (Section \ref{sec:action}). Again, \textit{NuGgeT} outperforms the baselines in all scenarios. Interestingly, the performance on WS and ER graphs seems to be much lower than for BA graphs. This can be understood empirically by analysing the eigenvalues of the normalised adjacency matrix $\mathbf{A}$ for different graphs. For WS and ER graphs, the smallest absolute (non-zero) eigenvalue of $\mathbf{A}$ is on average much smaller than for BA graphs (Fig.~\ref{fig:min_absolute_eigenvalues} in Appendix). As explained in section \ref{sec:action}, eigenvalues with very small absolute values will result in actions behaving like mid-spectrum eigenvectors (which are not necessarily smooth). We verify this further in Fig.~\ref{fig:graph_frequency_coefficients} in Appendix, which shows the spectral coefficients of the actions for all combinations of graphs and games. Linear Influence Games on ER and WS graphs are indeed the only scenarios where we empirically observe large graph Fourier coefficients for mid-spectrum eigenvectors. On BA graphs we do not observe this behaviour; mid-spectrum eigenvectors are most often not represented at all by the actions (the BA graphs we generate are trees that have zero eigenvalues associated with mid-spectrum eigenvectors which are discarded when taking the pseudoinverse), and actions tend to be smoother in this case. The results in Fig.~\ref{fig:linear_influence} suggest that when actions are less smooth (weaker association of actions with graph structure) all methods tend to perform less well.

\begin{figure}
    \vspace{-0.6cm}
    \centering
    \includegraphics[width=1\linewidth]{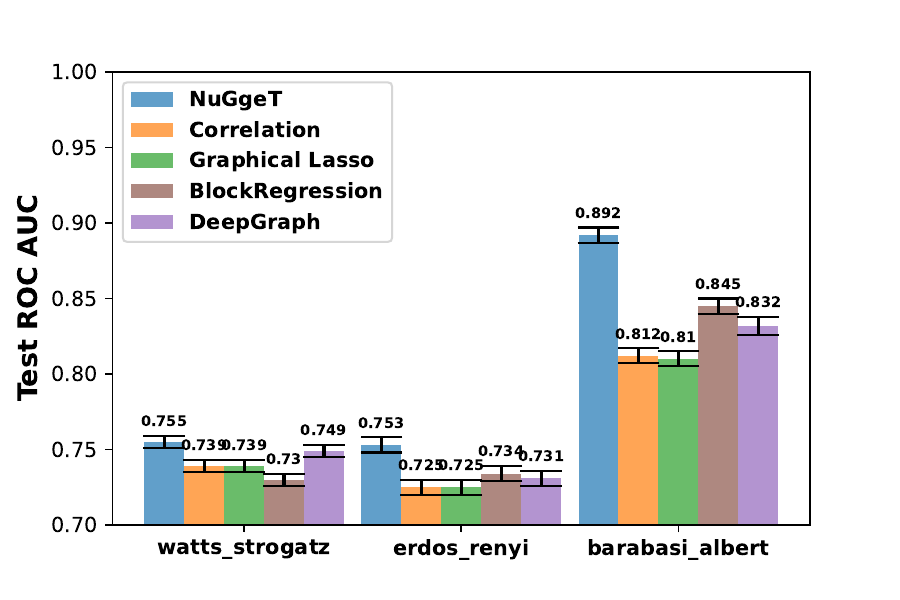}\vspace{-0.8cm}
    \caption{Results for BH  graphical games.}
    \label{fig:graphical_game}
\end{figure}

Results for BH Graphical Games are reported in Fig.~\ref{fig:graphical_game}. Since there are no parameters controlling the game, we only have 3 configurations corresponding to different graph types. Again, \textit{NuGgeT} outperforms other methods in all configurations, with the gap being largest on BA graphs. 
We also perform ablation studies on the number of games, number of training graphs and size of the graphs. The results are presented and discussed in Appendix.  

In many real-world scenarios, the observed actions will not be exactly at equilibrium but close to it, or trying to re-converge to the equilibrium after some perturbation. We therefore investigate how the performance of our model degrades with ``noisy'' samples where the sampled actions are only near equilibrium. Table \ref{tab:noise_results} shows the results for BA graphs and signals with unit norm. We can see performance remains satisfactory given reasonable amount of noise.

\begin{table}
\caption{Test accuracy on Barabási-Albert graphs for NuGgeT given different levels of Gaussian noise added to player actions.}
\vspace{0.2cm}
\scalebox{0.85}{
\begin{tabular}{|lccc|}
\hline
 Noise Std & Linear Quadratic & Linear Influence &   BH \\
\hline
  no noise &   99.87$\pm$0.02 &   99.87$\pm$0.02 &  99.87$\pm$0.02 \\
  0.10 &   99.33$\pm$0.10 &   99.33$\pm$0.10 &  99.33$\pm$0.10 \\
  0.20 &   95.50$\pm$0.38 &   95.50$\pm$0.38 &  95.50$\pm$0.38 \\
\hline
\end{tabular}}
\label{tab:noise_results}
\end{table}

To conclude, our method outperforms the baselines in all three types of games, most significantly 
in the scenarios where the actions are not necessarily smooth on the network and our method was able to infer such relationship from the observed data. While learning the mapping from actions to network, the model will learn not only the utility function of the games represented in the data, but also the typical structure of the networks. If all the networks in the training data are Barabási-Albert graphs, the prediction of the model will likely also be a Barabási-Albert graph. On the other hand, optimisation methods such as LinQuadOpt are not able to automatically bias their prediction to a particular class of graphs.

\subsection{Real-World Data}
Players in real-world scenarios often act according to strategic interactions.  It has been shown in the sociology and game theory literature that people have an incentive to conform to social norms~\citep{young2009innovation, montanari2010spread} or be influenced by their social network neighbours~\citep{leng2020learning}. That is, individual utilities are higher if their behaviours are similar to that of their neighbours in the social network. Such a mechanism will lead to strategic complement relationships. Following this assumption, we validate our model on two datasets, comparing with the baseline methods introduced in section \ref{sec:synthetic_data}.

\paragraph{Indian Villages}
The \textit{Indian Villages} dataset\footnote{The Indian Villages dataset can be accessed at \url{https://doi.org/10.7910/DVN/U3BIHX}.} \citep{banerjee2013diffusion} is a well-known dataset in the economics literature that contains data from a survey of social networks in 75 villages in rural southern Karnataka, a state in India. Each village constitutes a social network graph, where nodes are households and edges are self-reported friendships. Following the setup in \citet{leng2020learning}, we consider as actions the number of rooms, number of beds and other decisions families have to make related to their household. The reasoning is that if neighbours adopt a specific facility, villagers tend to gain higher payoff by doing the same, i.e., complying with social norms.
We only consider the 48 villages for which we have the ground truth actions and network. 40 are used for training, 3 for validation and 5 for testing. Categorical actions are one-hot encoded, while numerical actions are treated as continuous features. The resulting dataset has graphs with 10 actions and a number of nodes ranging between 77 and 356. It can be seen from Table \ref{tab:real_data_results} that \textit{NuGgeT} outperforms all other methods by at least $5.01\%$. 
\textit{DeepGraph} fails to learn altogether on this dataset.

\paragraph{Yelp Ratings}
The \textit{Yelp Ratings} dataset\footnote{The Yelp dataset can be accessed at \url{https://www.yelp.com/dataset}.} consists of rating of users to business, as well as the social connectivity between user. 
Similarly to the previous case, on deciding to review a local business, people may have an incentive to conform to the social norms they perceive, which are formed by the ratings from their neighbours. 
Yelpers tend to gain higher payoff with similar ratings due to social conformity, i.e., strategic complements in a game-theoretic context. 
From the raw data we extract 5000 sub-graphs representing communities, where the actions are the average rating of users to 22 categories of businesses. The task is to reconstruct the social connectivity of the users given their actions (ratings). 4250 graphs are used for training, 250 for validation and 500 for testing. More details of the dataset construction are provided in the appendix. On this dataset, NuGgeT outperforms all other baselines by at least $2.79\%$.
Overall, the results on real-world data show the efficacy of \textit{NuGgeT} in cases where the game utility is not explicitly known. 

\begin{table}[ht]
\footnotesize
    \centering
    \caption{Test ROC AUC for Indian Villages and Yelp Ratings data.}
    \vspace{0.2cm}
    \scalebox{0.85}{
        \begin{tabular}{| l  cc |}
        \hline
        Model                             & \texttt{Indian Villages}   & \texttt{Yelp Ratings}       \\ \hline
        \textit{Correlation}              & 0.5816$\pm$0.0135          & 0.6222$\pm$0.0043 \\
        \textit{Anticorrelation}          & 0.4184$\pm$0.0135          & 0.3778$\pm$0.0043 \\
        \textit{Graphical Lasso}          & 0.5823$\pm$0.0152          & 0.6523$\pm$0.0038 \\
        \textit{Baraki and Honorio}       & 0.5715$\pm$0.0164          & 0.6786$\pm$0.0032 \\ 
        \textit{LinQuadOpt (indep.)} & 0.5557$\pm$0.0108          & 0.6796$\pm$0.0033 \\ 
        \textit{LinQuadOpt (homop.)} & 0.5789$\pm$0.0174          & 0.6310$\pm$0.0036 \\ 
        \textit{DeepGraph}                & 0.4965$\pm$0.0143          & 0.6776$\pm$0.0039 \\ 
        \textit{NuGgeT}                   & \textbf{0.6324}$\pm$0.0167 & \textbf{0.7057}$\pm$0.0035 \\
        \hline
    \end{tabular}}
    \label{tab:real_data_results}
\end{table}

\section{Conclusion and Future Work} \label{sec:conclusion}
In this work, we propose a novel framework to infer the network structure behind the games from their equilibrium actions. Unlike existing methods, we achieve so by learning a mapping from the actions to the network structure without knowing the utility function of the game. This is especially beneficial in real-world scenarios where the nature of strategic interactions between players of the game remains hidden or may evolve over time.

\paragraph{Limitations and Future Work.} The current work only deals with static games and networks of small scale. Moreover, we do not deal with repeated games, i.e., where players have to take multiple actions sequentially. A promising future direction is therefore to extend it to dynamic games and networks, where both the utility function and the structure of the network may change over time, as well as dealing with larger graphs and repeated games. 
Moreover, we currently deal with test graphs that have a similar structure to the graphs the model has been trained on, e.g., we use the Barabási-Albert graphs during both training and testing. An interesting future work would be to work on the generalisation capability of this approach, i.e., the ability to be trained on one type of graph (e.g., Erdős-Rényi) but generalise to different types as well (e.g., Barabási-Albert or Watts-Strogatz).
In addition, our method cannot guarantee the uniqueness of the learned network; indeed, the main focus of the study is to propose a first and efficient data-driven learning framework without assuming the utility function and prior knowledge about the network structure. We leave the identification of the network structure, which is a challenging problem in itself, for future work. 

\section*{Acknowledgements}
Y.L. is supported by the NSF grant IIS-2153468. X.D. acknowledges support from the Oxford-Man Institute of Quantitative Finance and the EPSRC (EP/T023333/1).

\bibliography{reference}
\bibliographystyle{icml2022}

\newpage
\appendix
\onecolumn

\section{Proof of NuGgeT symmetries}
We would like to show that the \textit{NuGgeT} model $g$ is {\em permutation-invariant} over the set of games and {\em permutation-equivariant} over the set of nodes.

Starting with the first one, we need to show that $g(\mathbf{X} \mathbf{P}_2)=g(\mathbf{X})$, where $\mathbf{P}_2 $ is a $K\times K$ permutation matrix on the games. For what concerns the encoder, we only need to prove that $\alpha_{ij}^{(h)}$ is invariant w.r.t. permutation of the games. 
Let $p_k$ be the index of the non-zero entry of the $k$-th column of $P_2$, we have:

\begin{align*}
    \alpha_{ij}^{(h)} &= \mathop{\bigboxvoid}_{k=1}^K  {\mathbf{y}}_{ik}^\top \mathbf{W}_Q^{(h)}\mathbf{W}_K^{(h)}{\mathbf{y}}_{jk} \\
    &= \mathop{\bigboxvoid}_{k=1}^K  {\mathbf{y}}_{ip_k}^\top \mathbf{W}_Q^{(h)}\mathbf{W}_K^{(h)}{\mathbf{y}}_{jp_k},
\end{align*}

since $\mathop{\bigboxvoid}_{k=1}^K$ is chosen to be a permutation invariant operator. This shows that $\alpha_{ij}^{(h)}$ does not depend on the particular permutation $p$ chosen, i.e., it is invariant w.r.t. permutation of the games. Similarly, the same holds for the decoder as  $\mathop{\bigboxdot}_{k=1}^{K}$ is a permutation invariant operator and therefore $\mathop{\bigboxdot}_{k=1}^{K} \mathbf{z}_{ik} \odot \mathbf{z}_{jk} = \mathop{\bigboxdot}_{k=1}^{K} \mathbf{z}_{ip_k} \odot \mathbf{z}_{jp_k}$ for any permutation $p$.

Regarding the second part of the proof, we need to show that $g(\mathbf{P}_1 \mathbf{X})=\mathbf{P}_1 g(\mathbf{X})\mathbf{P}_1^\top$, where $\mathbf{P}_1$  is an $N\times N$ permutation matrix interpreted as reordering of the nodes of the graph. Letting $p_k$ be the index of the non-zero entry of the $k$-th row of $\mathbf{P}_1$. Note that:

\begin{align*}
    \mathrm{enc}(\mathbf{P}_1\mathbf{X})_{ik} 
    &= \psi\Big({\mathbf{y}}_{p_ik},  \sum_{j=1}^N \alpha_{p_ip_j}^{(1)} {\mathbf{y}}_{p_jk},  \ldots ,  \sum_{j=1}^N \alpha_{p_ip_j}^{(H)} {\mathbf{y}}_{p_jk} \Big) \\
    &= \psi\Big({\mathbf{y}}_{p_ik},  \sum_{j=1}^N \alpha_{p_ij}^{(1)} {\mathbf{y}}_{jk},  \ldots ,  \sum_{j=1}^N \alpha_{p_ij}^{(H)} {\mathbf{y}}_{jk} \Big) \\
    &= \mathbf{z}_{p_ik}
\end{align*}

where the second equality stems from the fact that summation is a permutation invariant operator (i.e., $\sum_{j=1}^N x_i = \sum_{j=1}^N x_{p_i}$ for any permutation $p$). We then have:

\begin{align*}
    \Big(\mathbf{P}_1\ g(\mathbf{X}) \mathbf{P}_1^\top\Big)_{ij} &= \hat{a}_{p_ip_j} \\
    &= \psi\Big(\mathop{\bigboxdot}_{k=1}^{K} \mathbf{z}_{p_ik} \odot \mathbf{z}_{p_jk}\Big) \\
    &= \psi\Big(\mathop{\bigboxdot}_{k=1}^{K} \mathrm{enc}(\mathbf{P}_1\mathbf{X})_{ik} \odot \mathrm{enc}(\mathbf{P}_1\mathbf{X})_{jk}\Big)\\ 
    &= g(\mathbf{P}_1 \mathbf{X})_{ij}.
\end{align*}

Since this holds for all indices $ij$, it follows that $g(\mathbf{P}_1 \mathbf{X})=\mathbf{P}_1 g(\mathbf{X})\mathbf{P}_1^\top$.

\section{Experiments}

\subsection{Synthetic Data Generation} \label{sec:synthetic_data_generation}
We generate the synthetic data from three different graph models: Erdős-Rényi (ER), Barabási-Albert (BA) and Watts-Strogatz (WS). In ER graphs, an edge is present with a probability of $p = 0.2$, independently from all other possible edges. In WS graphs, we set the exact degree of the nodes to be $k = log_2(N)$, with a probability of $p = 0.2$ for the random rewiring process. Finally, in BA graphs, nodes are added one at a time and each new node has $m=1$ edges which are preferentially attached to existing nodes with already high degree (this results in a tree graph). 

\subsection{Yelp Dataset Generation}
Starting from the raw data at https://www.yelp.com/dataset, we create a dataset by performing the following steps:

\begin{enumerate}
    \item We compute the rating for every user to every businesses category, by averaging the ratings a user has given to all business of each category
    \item We weight each edge in the original user-user graph with the fraction of common categories the two users rated at least once
    \item We cluster the above weighted graph using Graclus \cite{dhillon2007weightedg} with the objective of minimising the normalised cut. 
    \item We discard all the clusters with less than 10 nodes, or with very sparse ratings (less than 25\% of the categories with a rating for at least 25\% of the users). The result of this step is $\sim$27k graphs of different users.
    \item We rank the clusters extracted above by their density of ratings and keep the 5000 most dense clusters. Each cluster constitutes a graph associated with node attributes, which can be used to train and test our model.
\end{enumerate}

\subsection{Baselines}
In both the synthetic and real-world data experiments we compare with the following baselines:
\paragraph{Correlation} The Pearson correlation coefficient between the actions of two nodes. This works particularly well when the actions are homophilous over the graph, i.e., nodes connected in the graph tend to take similar actions. We implement this baseline ourselves. 
\paragraph{Anticorrelation} The negative of the Pearson correlation coefficient between the actions of two nodes. This works well in case of strategic substitutes, i.e., when nodes connected in the graph tend to take different actions. We implement this baseline ourselves. 
\paragraph{Graphical Lasso \cite{friedman_sparse_2008}} Computes a sparse penalised estimation of the inverse of the covariance matrix. We use SKGGM\footnote{https://github.com/skggm/skggm} QuicGraphicalLasso with empirical covariance initialisation for this. 
\paragraph{LinQuadOpt} Algorithm presented in \cite{leng2020learning} which assumes the form of the game to be linear quadratic. It has two version, one where the benefits are assumed to be independent (\textit{LinQuadOpt (Independent)}), and another for homophilous benefits (\textit{LinQuadOpt (Homophilous)}). We implement the algorithm ourselves (no public code was provided).
\paragraph{BlockRegression} Algorithm presented in \cite{barik2019provable}, which has been designed specifically for BH graphical games. We re-implement this baseline ourselves following the paper (no public code was provided).

\paragraph{DeepGraph} Algorithm presented in \cite{belilovsky2017learning} which recovers the graph from the covariance matrix using a series of dilated convolutions. We re-implement \textit{DeepGraph} in PyTorch ourselves taking as inspiration the public TensorFlow implementation of the authors.

All the baselines are tuned on the validation set. For both \textit{Graphical Lasso} and \textit{BlockRegression} we tune the regularisation parameter in the range ${10^k : k \in [-5, 5] }$ with an interval of one, whereas for \textit{LinQuadOpt} both regularisation parameters are tuned in ${10^k : k \in [-6, 1] }$, also with an interval of one. We train \textit{DeepGraph} using the same hyperparameters used for NuGget (Adam optimiser, learning rate of 0.001, batch size of 100 and a patience of 50), with $\lceil\log_2(N_{\text{max}})\rceil$ convolutional layers ($N_{\text{max}}$ corresponds here to the maximum number of nodes of any graph in the dataset) and dilation coefficient equal to $d_k = 2^{k - 1}$ for layer $k$ as specified in \cite{belilovsky2017learning}.

\subsection{NuGgeT Hyperparameters}
We do not perform any extensive hyperparameter tuning for the NuGgeT model, but instead use the same standard choice of hyperparameters (reported in Table \ref{tab:nugget_hyperparameters}) for all experiments.

\begin{table*}[h]
    \centering
        \begin{tabular}{| l  c |}
        \hline
        \textit{Hyperparameter Name}     & \textit{Value}    \\
        \hline
        $F$                              & 10                \\ 
        $F'$                             & 10                \\ 
        $H$ (num of heads)               & 10                \\ 
        $\psi$'s num layers              & 2                \\ 
        $\psi$'s hidden dim              & 100                \\ 
        \hline
    \end{tabular} 
    \caption{Hyperparameters used for \textit{NuGgeT} in all experiments.}
    \label{tab:nugget_hyperparameters}
\end{table*}

\section{Spectral Analysis of the Games}
\begin{figure}
    \centering
    \includegraphics[width=0.5\linewidth]{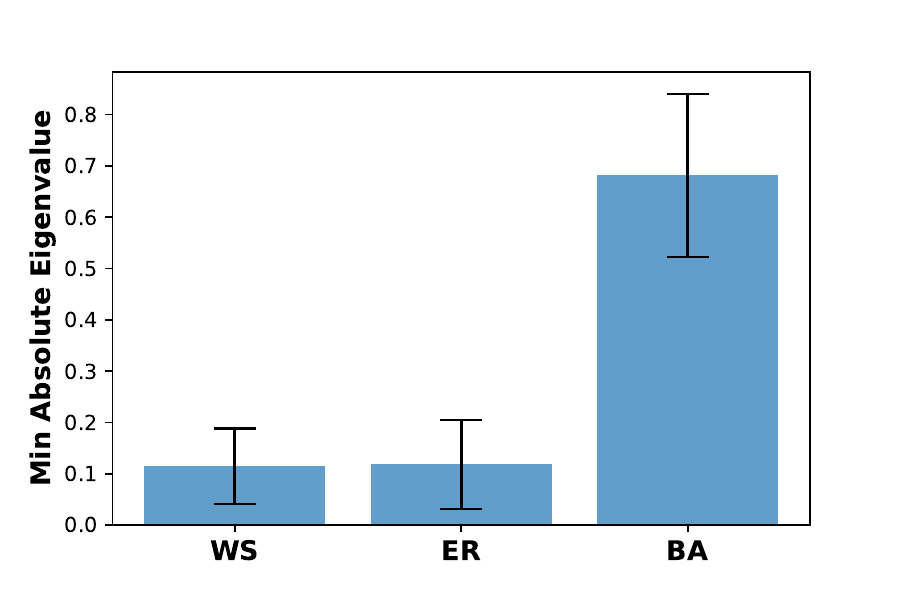}
    \caption{Mean and standard deviation of the minimum absolute non-zero eigenvalue of the adjacency matrix for different types of graph models. Statistics are computed over 1000 graphs with 20 nodes. }
    \label{fig:min_absolute_eigenvalues}
\end{figure}

\begin{figure}
    \centering
    \includegraphics[width=\linewidth]{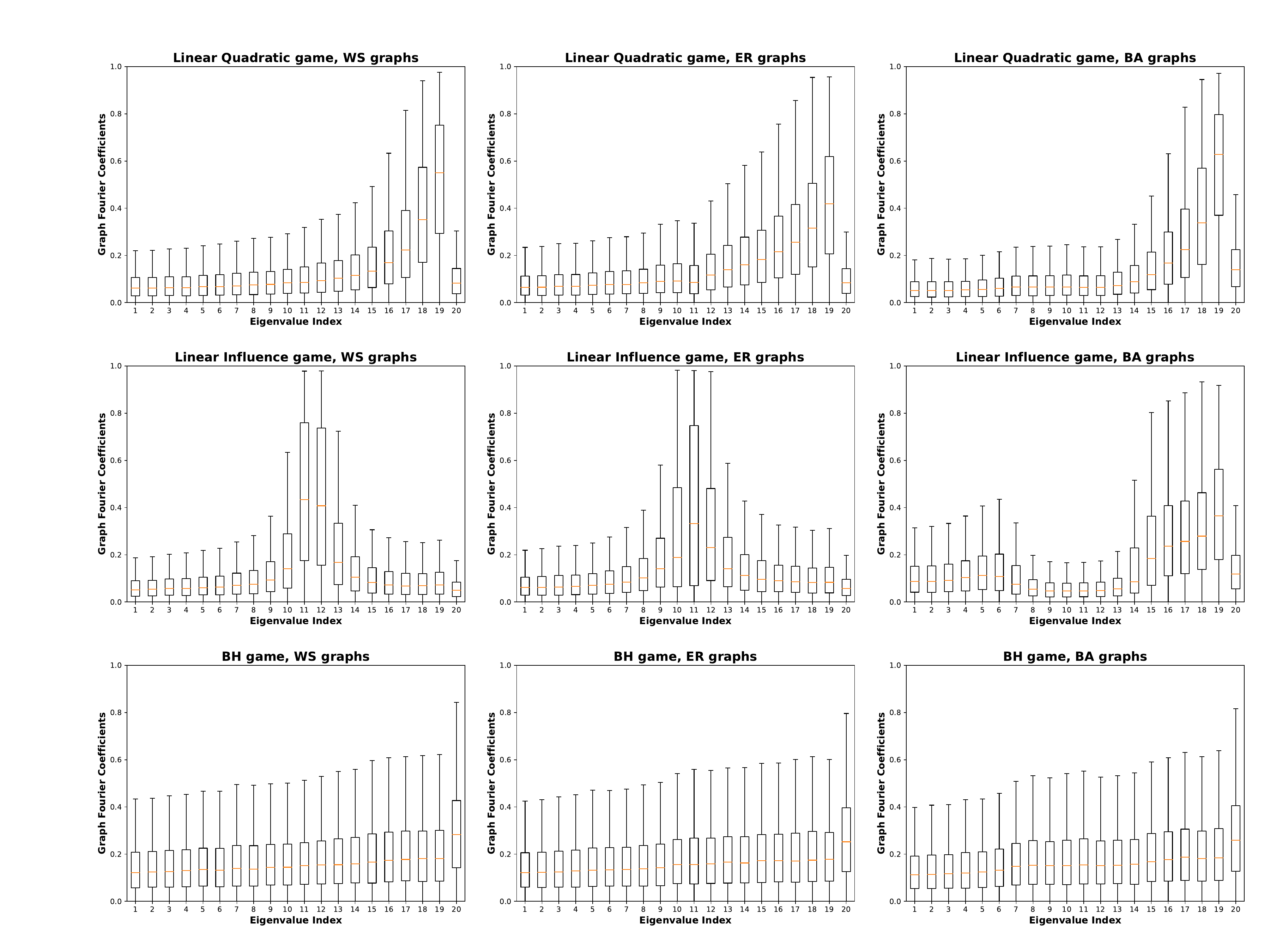}
    \caption{Magnitude of coefficients of normalised action vectors for different games and graphs. These are obtained by taking the Graph Fourier Transform of the actions, i.e., taking the inner product between the action and each of the adjacency matrix eigenvectors.}
    \label{fig:graph_frequency_coefficients}
\end{figure}

Fig.~\ref{fig:min_absolute_eigenvalues} shows that the smallest absolute (non-zero) eigenvalues of the normalised adjacency matrix $\mathbf{A}$ is on average much smaller for ER and WS graphs than for BA graphs. This results in the corresponding eigenvectors having a large influence on the actions for the Linear Influence games, where the Nash equilibrium actions satisfy  $\mathbf{x}^* = \mathbf{A}^{-1} \mathbf{b}$. This is confirmed by Fig.~\ref{fig:graph_frequency_coefficients}, which shows the graph Fourier coefficients for different eigenvalue indexes: the actions of linear influence games on ER and WS graphs are dominated by mid-range eigenvectors.

\section{Ablation Studies}
\begin{figure}
\vspace{-1cm}
    \centering
    \includegraphics[width=\linewidth]{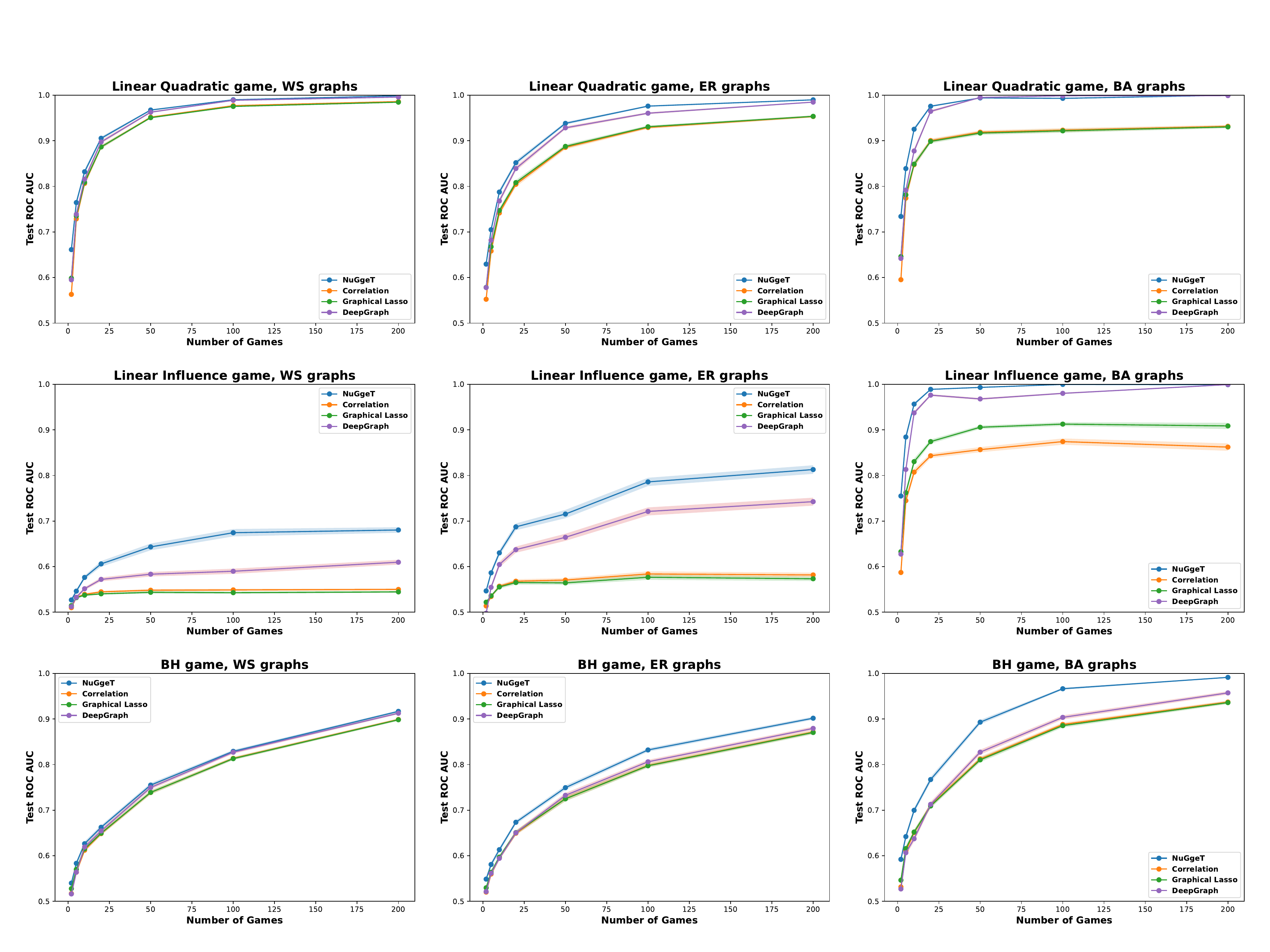}
    \caption{Results with varying number of games.}
    \label{fig:number_of_games}
\vspace{-0.5cm}
\end{figure}

\begin{figure}
    \centering
    \includegraphics[width=\linewidth]{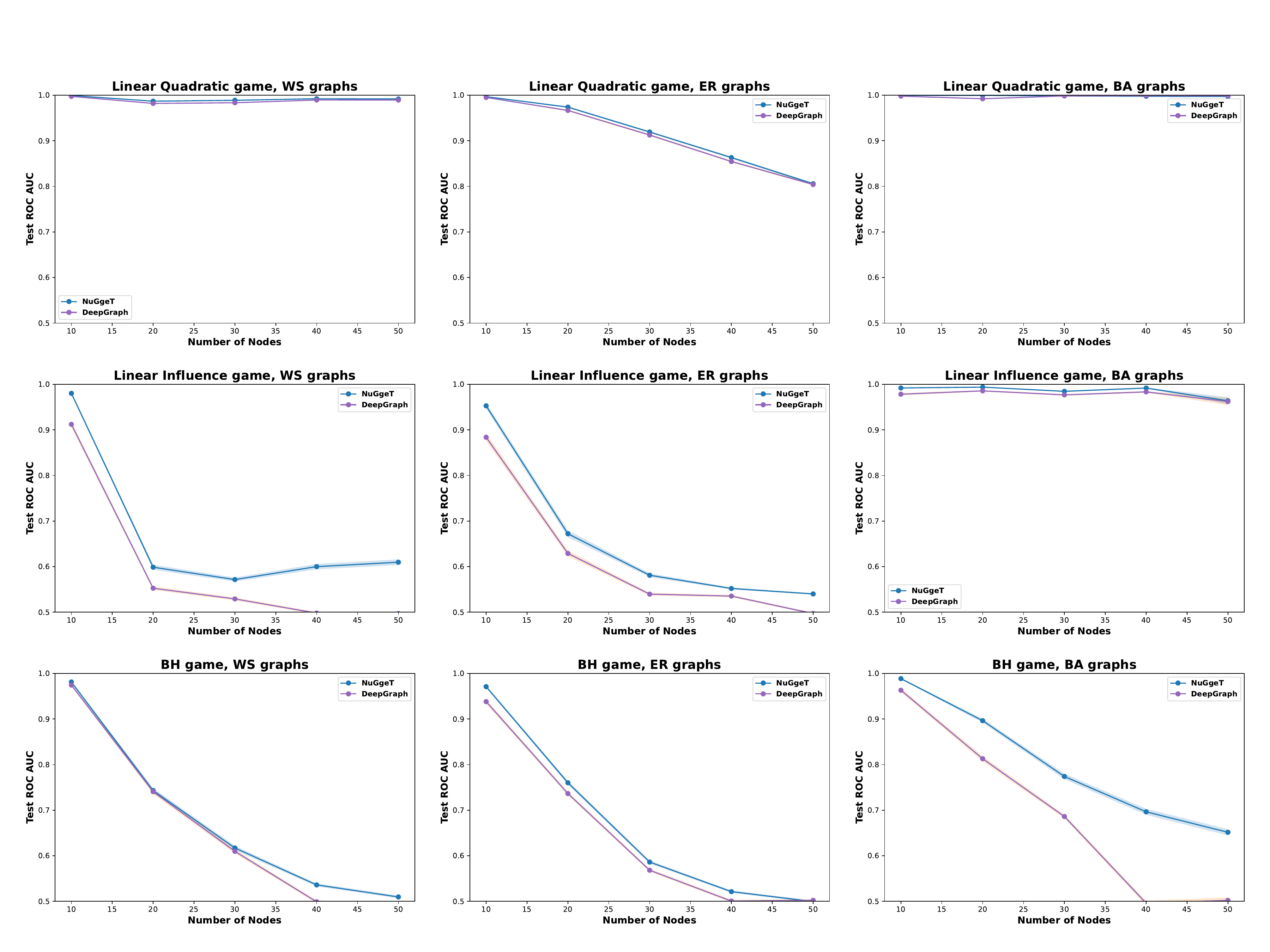}
    \caption{Results with varying number of nodes.}
    \label{fig:number_of_nodes}
\end{figure}

\begin{figure}
    \centering
    \includegraphics[width=\linewidth]{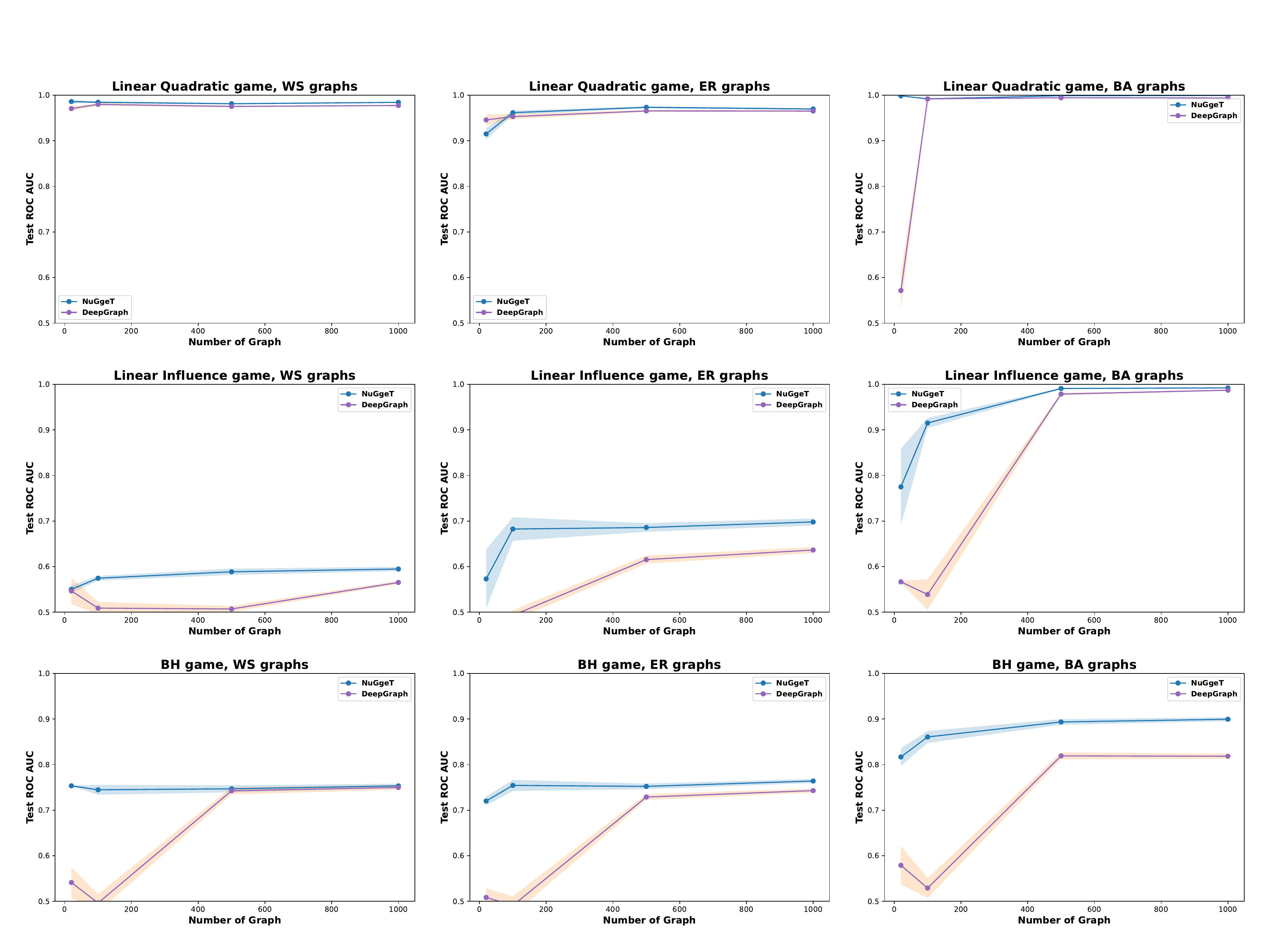}
    \caption{Results with varying number of training graphs.}
    \label{fig:number_of_graphs}
\end{figure}

In Fig.~\ref{fig:number_of_games}, we investigate how the performance changes with the number of games available. The smaller the number of games, the less information to reconstruct the graph from.
We use $\alpha=1$ for both linear quadratic and linear influence games, and a spectral radius of $0.6$ for linear quadratic games. In line with our expectations, all methods generally improve as more games are available.
We analyse the effect of larger graph sizes (Fig.~\ref{fig:number_of_nodes}). The more nodes, the more edge combinations exist and the harder the task becomes, which explains the decrease in performance of both methods as the number of nodes increases. Interestingly, the magnitude of the drop depends heavily on the combination of game and graph types, but \textit{NugGeT} seems to be more robust than \textit{DeepGraph}. We also investigate the effect of the number on training graphs on the model performance (Fig.~\ref{fig:number_of_graphs}). \textit{NugGeT} requires less training graphs compared to \textit{DeepGraph} to obtain a similar performance.
Fig.~\ref{fig:number_of_games} and \ref{fig:number_of_nodes} show respectively the performance of various methods when varying the number of games and the number of nodes respectively. We use $\alpha=1$ for both linear quadratic and linear influence games, and $\beta=0.6$ (strategic complements) for linear quadratic games. In line with our expectations, all methods generally improve as more games are available. On the other hand, the more nodes, the more edge combinations exist and the harder the task becomes, which explains the decrease in performance of all methods as the number of nodes increases. Interestingly, the magnitude of the drop depends heavily on the combination of game and graph types, but \textit{NugGeT} seems to be more robust than \textit{DeepGraph}. 

\section{\textit{NuGgeT}'s Attention Mechanism}

\begin{figure}[!ht]
    \centering
    
    \begin{minipage}{0.5\linewidth}
        \centering
        \scalebox{0.68}{
\begin{tikzpicture}

\begin{axis}[
colorbar,
colorbar style={ylabel={}},
colormap/viridis,
point meta max=1,
point meta min=0,
tick align=outside,
tick pos=left,
title={Attention on scalar actions ($a > 0$)},
x grid style={white!69.0196078431373!black},
xlabel={$x_j$},
xmin=-0.5, xmax=200.5,
xtick style={color=black},
xtick={0,40,80,120,160,200},
xticklabels={-1.0,-0.6,-0.2,0.2,0.6,1.0},
y dir=reverse,
y grid style={white!69.0196078431373!black},
ylabel={$x_i$},
ymin=-0.5, ymax=200.5,
ytick style={color=black},
ytick={0,40,80,120,160,200},
yticklabels={-1.0,-0.6,-0.2,0.2,0.6,1.0}
]
\addplot graphics [includegraphics cmd=\pgfimage,xmin=-0.5, xmax=200.5, ymin=200.5, ymax=-0.5] {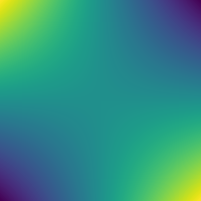};
\end{axis}

\end{tikzpicture}
        }
    \end{minipage}\hfill
    \begin{minipage}{0.5\linewidth}
        \centering
        \scalebox{0.68}{
\begin{tikzpicture}

\begin{axis}[
colorbar,
colorbar style={ylabel={}},
colormap/viridis,
point meta max=1,
point meta min=0,
tick align=outside,
tick pos=left,
title={Attention on scalar actions ($a < 0$)},
x grid style={white!69.0196078431373!black},
xlabel={$x_j$},
xmin=-0.5, xmax=200.5,
xtick style={color=black},
xtick={0,40,80,120,160,200},
xticklabels={-1.0,-0.6,-0.2,0.2,0.6,1.0},
y dir=reverse,
y grid style={white!69.0196078431373!black},
ylabel={$x_i$},
ymin=-0.5, ymax=200.5,
ytick style={color=black},
ytick={0,40,80,120,160,200},
yticklabels={-1.0,-0.6,-0.2,0.2,0.6,1.0}
]
\addplot graphics [includegraphics cmd=\pgfimage,xmin=-0.5, xmax=200.5, ymin=200.5, ymax=-0.5] {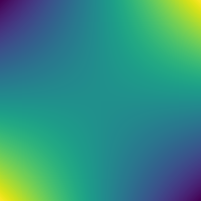};
\end{axis}

\end{tikzpicture}
        }
    \end{minipage}
    \begin{minipage}{0.5\linewidth}
        \centering
        \scalebox{0.68}{
\begin{tikzpicture}

\begin{axis}[
colorbar,
colorbar style={ylabel={}},
colormap/viridis,
point meta max=1,
point meta min=0,
tick align=outside,
tick pos=left,
title={Attention on expanded actions},
x grid style={white!69.0196078431373!black},
xlabel={$x_j$},
xmin=-0.5, xmax=200.5,
xtick style={color=black},
xtick={0,40,80,120,160,200},
xticklabels={-1.0,-0.6,-0.2,0.2,0.6,1.0},
y dir=reverse,
y grid style={white!69.0196078431373!black},
ylabel={$x_i$},
ymin=-0.5, ymax=200.5,
ytick style={color=black},
ytick={0,40,80,120,160,200},
yticklabels={-1.0,-0.6,-0.2,0.2,0.6,1.0}
]
\addplot graphics [includegraphics cmd=\pgfimage,xmin=-0.5, xmax=200.5, ymin=200.5, ymax=-0.5] {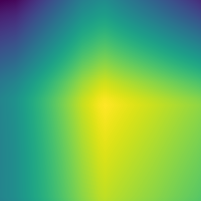};
\end{axis}

\end{tikzpicture}
        }
    \end{minipage}\hfill
    \begin{minipage}{0.5\linewidth}
        \centering
        \scalebox{0.68}{
\begin{tikzpicture}

\begin{axis}[
colorbar,
colorbar style={ylabel={}},
colormap/viridis,
point meta max=1,
point meta min=0,
tick align=outside,
tick pos=left,
title={Attention on expanded actions},
x grid style={white!69.0196078431373!black},
xlabel={$x_j$},
xmin=-0.5, xmax=200.5,
xtick style={color=black},
xtick={0,40,80,120,160,200},
xticklabels={-1.0,-0.6,-0.2,0.2,0.6,1.0},
y dir=reverse,
y grid style={white!69.0196078431373!black},
ylabel={$x_i$},
ymin=-0.5, ymax=200.5,
ytick style={color=black},
ytick={0,40,80,120,160,200},
yticklabels={-1.0,-0.6,-0.2,0.2,0.6,1.0}
]
\addplot graphics [includegraphics cmd=\pgfimage,xmin=-0.5, xmax=200.5, ymin=200.5, ymax=-0.5] {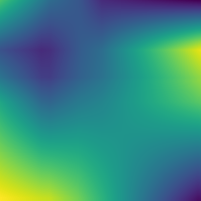};
\end{axis}

\end{tikzpicture}
        }
    \end{minipage}
    \caption{Top: attention values for a multiplicative attention layer directly operating on the input scalars $x_i \in [-1, 1]$, $x_j \in [-1, 1]$ with positive (left) and negative (right) coefficient. Bottom: attention values for a multiplicative attention layer with random coefficients processing the expanded actions $\mathbf{y}_i$, $\mathbf{y}_j$ of $x_i$, $x_j$. The result is scaled to $[0, 1]$ for visualisation purposes.}
    \label{fig:attention_scores}
\end{figure}

We investigate the difference between a multiplicative attention mechanism operating directly on scalar actions (i.e., with unnormalised attention coefficients of the form: $\tilde{\alpha}_{ij} = x_i x_j a$ with $a$ a learnable parameter), and one operating on transformed actions (i.e., $\tilde{\alpha}_{ij} = \mathbf{y}_{i}^\top \mathbf{W}_Q\mathbf{W}_K\mathbf{y}_{j}$, with $\mathbf{y}_{i}      = \text{ReLU}(x_{i} \boldsymbol{w} + \mathbf{b})$).
In Fig.~\ref{fig:attention_scores} we plot the non-normalised attention values of the two different mechanisms for varying values of the input actions $x_i$, $x_j$. We use a positive and a negative value of $a$ for showing the capability of the simple scalar-based attention mechanism to identify correlation and anti-correlation in the input actions $x_{.}$ and random coefficients drawn from a normal distribution for $\mathbf{W}_Q$, $\mathbf{W}_K$, $\boldsymbol{w}$, $\mathbf{b}$ for the attention mechanism used in \textit{NuGgeT}. While the scalar attention always shows a linear trend for a fixed $x_i$ (or $x_j$), the multiplicative layer operating on vectors is able to produce richer non-linear attention patterns with maximum / minimum values for the interior points of $[-1, 1]$.

You can have as much text here as you want. The main body must be at most $8$ pages long.
For the final version, one more page can be added.
If you want, you can use an appendix like this one, even using the one-column format.

\end{document}